\documentclass{article}

\usepackage{arxiv}
\usepackage[utf8]{inputenc} % allow utf-8 input
\usepackage[T1]{fontenc}    % use 8-bit T1 fonts
\usepackage{hyperref}       % hyperlinks
\usepackage{url}            % simple URL typesetting
\usepackage{booktabs}       % professional-quality tables
\usepackage{amsfonts}       % blackboard math symbols
\usepackage{nicefrac}       % compact symbols for 1/2, etc.
\usepackage{microtype}      
\usepackage{lipsum}		% Can be removed after putting your text content
\usepackage{graphicx}
\usepackage{doi}
\usepackage{standalone}
\usepackage[short]{optidef}
\usepackage{multirow}
\usepackage[justification=centering]{caption}
\usepackage{subcaption}
\usepackage{bm}
\usepackage{diagbox}
\usepackage{lipsum}
\usepackage[table]{xcolor}
\usepackage{amssymb}
\usepackage{amsthm}
\usepackage{mathcommands}
\usepackage{standalone}
\usepackage{amsmath}
\theoremstyle{definition}
\newtheorem{definition}{Definition}
\title{Advances in the training, pruning  and  enforcement  of  shape  constraints  of  Morphological Neural Networks using Tropical Algebra}
\newcommand\blfootnote[1]{%
  \begingroup
  \renewcommand\thefootnote{}\footnote{#1}%
  \addtocounter{footnote}{-1}%
  \endgroup
}
\author{ Dimitriadis Nikolaos \\
	\'{E}cole Polytechnique F\'{e}d\'{e}rale de Lausanne \\
	Lausanne Switzerland \\
	\texttt{nikolaos.dimitriadis@epfl.ch} \\
	\And
	Petros Maragos \\
	National Technical University of Athens \\
	Athens, Greece \\
	\texttt{maragos@cs.ntua.gr} \\
}

\renewcommand{\headeright}{}
\renewcommand{\undertitle}{}
\renewcommand{\shorttitle}{\textit{arXiv}}

\hypersetup{
    colorlinks=true,
    linkcolor=blue,
    filecolor=blue,      
    urlcolor=blue,
    citecolor=blue
}
\newcommand{\g}{\cellcolor{green}}
\newcommand{\bbmm}{\cellcolor{red!10}}
\newcommand{\bb}{\cellcolor{red!40}}

\usepackage[
    backend=biber,
    style=numeric,
    sorting=none,
    giveninits=true,
    maxcitenames=2
]{biblatex}
\DeclareNameAlias{author}{family-given}

\DeclareMathOperator{\argmax}{argmax}

\newcommand{\maxmatrmult}{\ensuremath{\boxplus}} % max-plus matrix multiplication 
\newcommand{\minmatrmult}{\ensuremath{\boxplus'}} % min-plus matrix multiplication
\def\rn{\mathbb{R}^{n}  }

\newcommand\numberthis{\addtocounter{equation}{1}\tag{\theequation}}
\newcommand{\tagg}[1]{\tag*{\textcolor{blue}{\( \triangleright \) {\small #1} }}}

\begin{document}
	\maketitle
	\blfootnote{\( ^* \) This work was performed when N.Dimitriadis was at the 
	National Technical University of Athens.}
	\begin{abstract}
    In this paper we study an emerging class of neural networks based on the morphological operators of dilation and erosion. We explore these networks mathematically from a tropical geometry perspective as well as mathematical morphology. Our contributions are threefold. First, we examine the training of morphological networks via Difference-of-Convex programming methods and extend a binary morphological classifier to multiclass tasks. Second, we focus on the sparsity of dense morphological networks trained via gradient descent algorithms and compare their performance to their linear counterparts under heavy pruning, showing that the morphological networks cope far better and are characterized with superior compression capabilities. Our approach incorporates the effect of the training optimizer used and offers quantitative and qualitative explanations. Finally, we study how the architectural structure of a morphological network can affect shape constraints, focusing on monotonicity. Via Maslov Dequantization, we obtain a softened version of a known architecture and show how this approach can improve training convergence and performance.
\end{abstract}


	\section{Introduction}
\label{sec:intro}

In the past decade, the field of neural networks has garnered research interest in the machine learning community, paving the way for the formation of a novel field called \textit{Deep Learning}. The cell of the models is the neuron, introduced by Rosenblatt. The neuron mimics the transformations of data performed in biological organisms. In mathematical terms, it consists of a multiply-accumulate scheme that is succeeded by a nonlinearity, called \textit{activation function}. An alternative lies in morphology-based models. 

In morphological neural networks the operations of addition and multiplication of the aforementioned multiply-accumulate scheme are replaced by maximum (or minimum) and addition, respectively. This process is called \textit{tropicalization} and yields a path towards tropical mathematics. An important aspect of the operator change is the lack of need for an activation function, since maximum (or minimum) are inherently nonlinear operations. Networks with this modified neuron have been studied in the context of neural networks \cite{ritter1996introduction,ritter1998morphological,sussner1998morphological,pessoa2000neural,yang1995min}. 
Recently, the field of tropical geometry has been linked with this class of morphological networks \cite{Charisopoulos_Maragos_2017,Charisopoulos_Maragos_2018,zhang2019max,SmyrnisMaragos_2020_ICML,SmyrnisMaragos_2020_ICML}. Tropical geometry studies {\it piecewise linear (PWL)} surfaces whose arithmetic is governed by a tropical semiring, where ordinary addition is replaced by the maximum or minimum and ordinary multiplication is replaced by ordinary addition. We refer to these algebraic structures as \( (\max,+) \) and \( (\min,+) \) semirings, respectively. These two semirings are dual and linked via the isomorphism $\phi(x)=-x$. 

In this paper, we study morphological networks via mathematical constructs stemming from mathematical morphology, lattice theory and tropical algebra. Our contributions are multifaceted and can be summarized as follows:

\begin{itemize}
    \item An alternative training process for morphological networks for general multiclass classification tasks is presented. By formulating the classification problem as an instance of a Difference-of-Convex optimization problem, the training can be performed with analytical and robust methods \textit{not} based on stochastic optimization.
    \item We explore the compression abilities of morphological networks and draw favorable conclusions in comparison with their linear counterparts. We show that morphological networks encode information with fewer parameters.
    \item We present a method to enforce monotonicity on the output of a morphological network using smooth operator approximations and show how the softened versions improve performance and training convergence.
\end{itemize}

The structure of the paper is the following. 
Section \ref{sec:related work} reviews past work that motivate our findings.
 In Section \ref{sec:background} we introduce some mathematical background concepts that accompany the ideas of the next sections and highlight the connections of mathematical morphology and tropical geometry. In Section \ref{sec:ccp training} we introduce an alternative method of morphological neural network training based on a heuristic called \textit{Convex-Concave Procedure} and proposed by \citeauthor[]{Charisopoulos_Maragos_2017} \cite{Charisopoulos_Maragos_2017} and propose an extension to general (i.e. multiclass) classification problems. In Section \ref{sec:sparsity} we study denser morphological networks and shift our focus from accuracy to sparsity. We use a pruning method to evaluate how the removal of parameters affects performance on both morphological networks and their linear counterparts concluding with the former's superiority in that regard. Section \ref{sec:monotonicity} explores regression with monotonicity constraints and focuses on neural network methods, exploring the smoothening of the "hard" morphological operators \( \min \) and \( \max \) to achieve better performance and simultaneously alleviate some training issues. Section \ref{sec:conclusion} includes concluding remarks.

	\section{Related Work}
\label{sec:related work}

\citeauthor[]{davidson1993morphology} introduced the morphological neuron \cite{davidson1993morphology}, aiming at learning morphological elements such as erosion and dilation. These efforts were intensified and led to more general models, such as a simple network with a hidden layer tasked to solve binary classification problems \cite[]{ritter1996introduction}. These models use different operators than addition and multiplication and are rooted in Mathematical Morphology and Lattice Theory \cite{Maragos_2017}. The model proposed by \citeauthor{ritter1996introduction} \cite[]{ritter1996introduction} constructs a decision boundary parallel to the axes, a limitation that has been addressed in two major ways. First, the architecture was extended to include a second hidden layer \cite{sussner1998morphological}. This modification allows the network to learn multiple axis-parallel decision boundaries and, thus, tackle more general problems. Second, \citeauthor{barmpoutis2007orthonormal} resolved this limitation by rotating the hyperplanes \cite{barmpoutis2007orthonormal}. 

\citeauthor{ritter1998morphological} introduced the term \textit{morphological networks} by replacing addition and multiplication with maximum and addition \cite[]{ritter1998morphological}. They focused on the storage and computing capabilities of associative memories based on the morphological neuron. 
\citeauthor{ritter2003lattice} link the morphological neuron with biological processes, introduce its dendritic structure and show that morphological perceptrons with a single hidden layer can approximate any high dimensional compact region within an arbitrary degree of accuracy \cite{ritter2003lattice}. Furthermore, \citeauthor{sussner2011morphological} study such networks in the prism of competitive learning, where the operator \( \argmax \) is employed to create a winner-take-all strategy at the output layer \cite{sussner2011morphological}.

\citeauthor{yang1995min} introduced the class of \textit{min-max} classifiers which consist of two layers, one with \( \max \) terms and one with \( \min \) terms. The models are trained within the context of Probably Approximately Correct (PAC) learning and can be considered as a generalization of Boolean functions based on Lattice Theory \cite{yang1995min}.  \citeauthor{pessoa2000neural} proposed a hybrid neuron that combines morphological and linear terms \cite{pessoa2000neural}.
A common theme in the morphological network literature is the formulation of gradient descent variants for training to address the issue of non-differentiability of morphological operators.

Using the tropical mathematics framework, \citeauthor{Charisopoulos_Maragos_2017} studied these networks and proposed a training algorithm not rooted in stochastic optimization but rather on Difference-of-Convex Programming \cite{Charisopoulos_Maragos_2017} and provided lower bounds for the linear regions of such networks \cite{Charisopoulos_Maragos_2018}. \citeauthor{Zhang_Naitzat_Lim_2018} study feedforward network with Rectified Linear Unit (ReLU) activations and showed their equivalence with tropical rational maps \cite{Zhang_Naitzat_Lim_2018}. Also, \citeauthor{Calafiore_Gaubert_Possieri_2018} study a similar class of networks with \textit{Log-Sum-Exp} terms, which can be thought of as smooth approximations of the hard morphological operators \cite{Calafiore_Gaubert_Possieri_2018,Calafiore_Gaubert_Member_Possieri_2019}. 
Tropical mathematics have also been linked with sparse systems and the problem of pruning models. \cite{Tsiamis_Maragos_2019} studies the sparsity of max-plus systems. Other works have focused on neural network pruning; \citeauthor{smyrnis2020maxpolynomial} develop a geometric algorithm for tropical polynomial division aimed at neural network minimization \cite{smyrnis2020maxpolynomial} and extend this method to multiclass problems \cite{SmyrnisMaragos_2020_ICML}.

In recent years, the field of morphological networks has adopted the trend of Deep Learning with increasingly more complex architectures, both in terms of width as well as depth. \citeauthor[]{Mondal_Santra_Chanda_2019} study dense morphological networks by choosing the morphological operators of dilation and erosion as the basic operations of a neuron and overcome their undifferentiability by considering smooth approximations \cite{Mondal_Santra_Chanda_2019}. Other works have focused on extending the morphological setting to convolutional layers. \citeauthor{franchi2020deep} proposed a joint operation that includes pooling and the morphological non-linearities \cite{franchi2020deep}, while \citeauthor{mellouli2019morphological} employed the counter-harmonic mean to inject morphological operations to convolutional layers and created an interpretable morphological convolutional neural network aimed at digit recognition \cite{mellouli2019morphological}.

Morphological Neural Networks have also been proposed to address monotonicity constraints. First, in the context of neural networks, \citeauthor{Archer_Wang_1993} proposed to a heuristic which updates the weights of a binary classification model so that samples do not violate the monotonicity constraints. The resulting network has only positive weights (for the case of increasing monotonicity) \cite{Archer_Wang_1993}. A different approach lies in constraining the weights to positive values in a neural network with a single hidden layer. This can be achieved through a monotonic nonlinear transformation with a positive images such as the sigmoid function \cite{kay2000estimating}. However, \citeauthor{Velikova_Daniels_Feelders_2006} showed that this approach requires \( K \) hidden layers to approximate a \( K \)-dimensional monotonic surface \cite{Velikova_Daniels_Feelders_2006}. On the other hand \citeauthor{Sill} proposes a network that guarantees monotonicity for the output via architectural design \cite[]{Sill}. This method is extended by \citeauthor{Daniels_Velikova_2010} to include cases of partially monotone functions \cite[]{Daniels_Velikova_2010}. Also, Sill's work can also be considered a special case of the min-max classifiers proposed by \citeauthor{yang1995min} \cite[]{yang1995min}.

Apart from morphological networks, other monotonic methods have been proposed in the machine learning field. An overview appears in \cite{gupta2016monotonic}. Some efforts have been focused on interpolated look-up tables, where shape constraints including monotonicity can be imposed \cite{Gupta_Bahri_Cotter_Canini_2018,Milani_Fard_Canini_Cotter_Pfeifer_Gupta_2016} and these ideas have also been applied in the context of deep networks \cite{You_Ding_Canini_Pfeifer_Gupta_2017}. An alternative approach has been proposed by \citeauthor{Wehenkel_Louppe_2019}, where monotonic functions are modeled via neural network techniques by enforcing a positive image on the derivative of the monotonic function \cite{Wehenkel_Louppe_2019}.

	\section{BACKGROUND CONCEPTS} 
\label{sec:background} 

\begin{figure}[hptb!]
    \centering
    \includegraphics[width=0.6\linewidth]{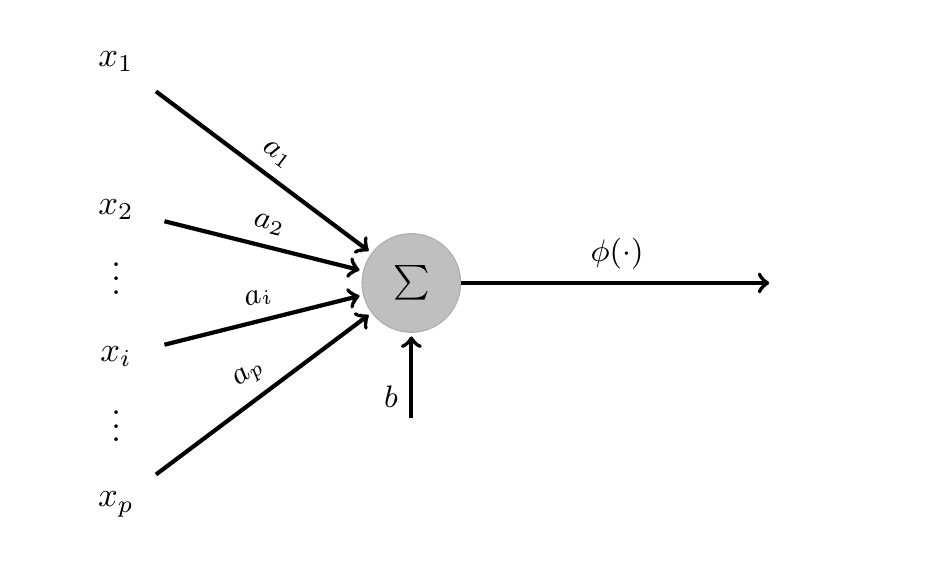}
    \caption{Perceptron}
    \label{fig:perceptron}
\end{figure}

In machine learning literature, the perceptron uses a multiply-accumulate scheme that feeds into an activation function \( \phi(\cdot) \), which performs a nonlinear transformation. A graphical representation appears in Fig. \ref{fig:perceptron}.  Specifically, the perceptron consists of a weight vector \( \mathbf{a}\in\rn \) and a bias term \( b\in\r \). These parameters are combined with the input \( \x\in\rn \) to compute the weighted sum \( f(\x)=\mathbf{a}^\top \x+b=\sum_{i=1}^{n} a_ix_i+b  \), which then passes from the aforementioned nonlinear filter. The tropical or morphological neuron has similar structure but performs different operations. The operations of addition and multiplication are replaced by \( \max \) (or \( \min \)) and addition. This process is known as \textit{tropicalization}. A notable difference is the absence of a nonlinear activation function, since the operators \( \max \) (or \( \min \) are inherently nonlinear. In the context of tropical algebra, we use the symbols \( \maxmatrmult \) and \( \minmatrmult \) to denote max-plus and min-plus matrix multiplication, respectively. These operations are defined as \( (\mathbf{ A }\maxmatrmult \mathbf{ B })_{ij}=\bigvee_{q=1}^k a_{iq}+b_{qj} \) and \( (\mathbf{ A }\minmatrmult \mathbf{ B })_{ij}=\bigwedge_{q=1}^k a_{iq}+b_{qj} \) for matrices of appropriate dimensions.

\begin{figure}[h]
    \def\ll{0.9}
    \def\between{0.48}
    \centering
    \begin{subfigure}[b]{\between\textwidth}
        \centering
        \includegraphics[width=\ll\linewidth]{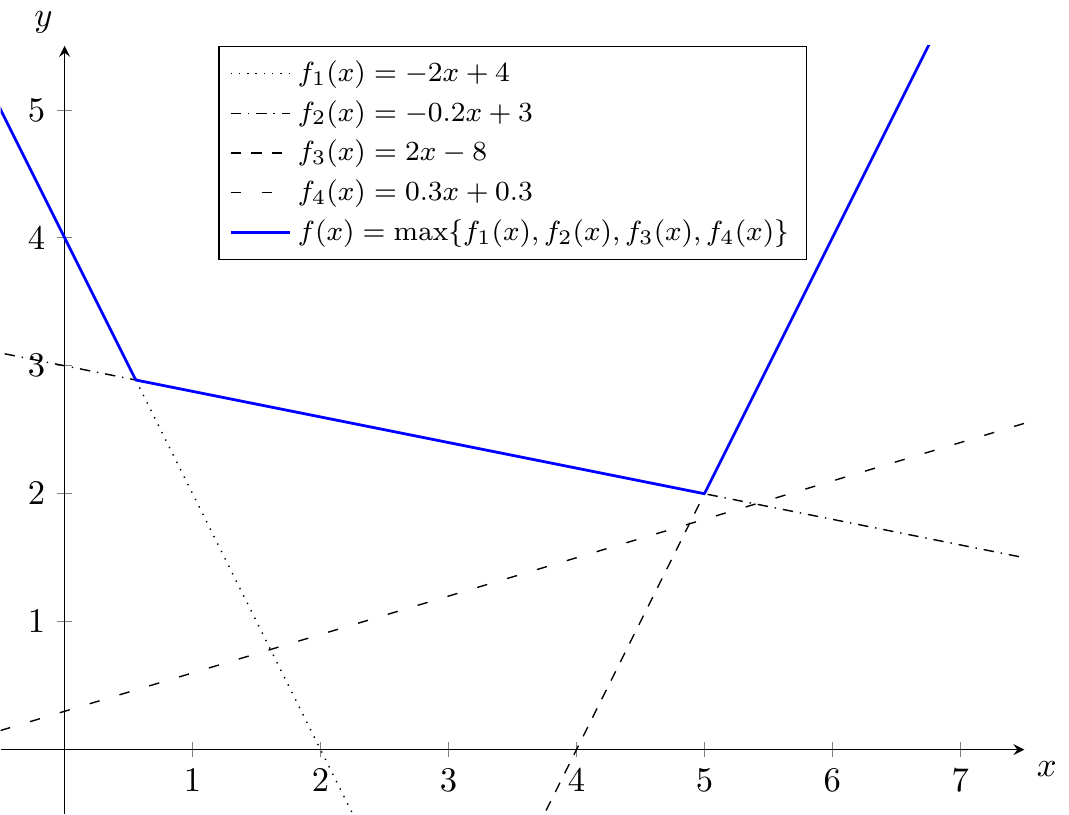}
        % \includestandalone[width=\ll\linewidth,mode=image]{media/plots/polynomial_maxplus}
        \caption{max-plus polynomial}
        \label{fig:polynomial_maxplus}
    \end{subfigure}
    \hfill
    \begin{subfigure}[b]{\between\textwidth}
        \centering
        \includegraphics[width=\ll\linewidth]{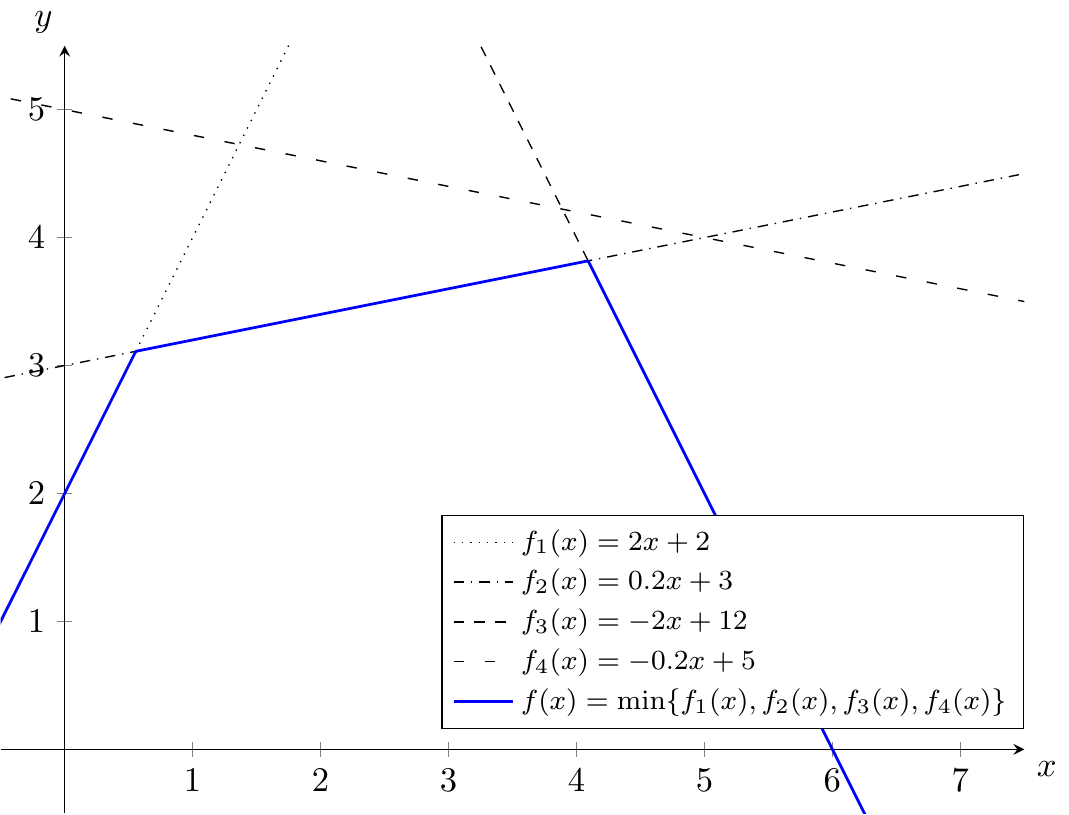}
        \caption{min-plus polynomial}
        \label{fig:polynomial_min-plus}
    \end{subfigure}
    \caption{Constructing polynomials in the tropical semiring. In both cases, \( f_4(x) \) has no bearing in the resulting surface. This term can be removed.}
    \label{fig:maxplus and minplus polynomials}
\end{figure}

The operator change yields piecewise linear (PWL) surfaces. For \( \mathbf{x}\in\rn \), the tropical max-plus polynomial \( p_\vee \) is defined as the maximum of multiple affine terms:
\begin{equation}
    \label{eq:tropical polynomial}
    p_\vee(\x)=\max_i\{\mathbf{a}_i ^\top \x+b_i\} = \bigvee_i \mathbf{a}_i ^\top \x+b_i,
\end{equation}
whilst its min-plus equivalent arises from replacing \( \max \) with \( \min \). Examples of max-plus and min-plus polynomials are presented in Fig. \ref{fig:maxplus and minplus polynomials}. The max-plus (min-plus) polynomial generates convex (concave) PWL surfaces. In the context of mathematical morphology, these operators are often referred to as dilation \( \dilation \) and erosion \( \erosion \), respectively, and are defined as:
\begin{align}
    \label{eq:dilation}
    \dilation_\mathbf{w}(\mathbf{x})&=w_0\vee\left(\bigvee w_i+x_i\right) \\
    \label{eq:erosion}
    \erosion_\mathbf{m}(\mathbf{x})&=m_0\wedge\left(\bigwedge m_i+x_i\right)
\end{align}
where \( \mathbf{w,m} \) correspond to the weights of the dilation and the erosion neurons, respectively. These expressions are special cases of tropical polynomials, since the term \( \mathbf{a}_i \) is specific and can be expressed as tropical matrix multiplications as \( \dilation_\mathbf{w}(\mathbf{x})=\mathbf{w} ^\top \maxmatrmult \mathbf{\tilde{x}} \) and \( \erosion_\mathbf{m}(\x)=\mathbf{m}^\top \minmatrmult \mathbf{\tilde{x}} \) where \( \mathbf{\tilde{x}}=\begin{bmatrix}
    1 & \x
\end{bmatrix} \), i.e. includes the bias term. 
Dilation and erosion are commonly used in image processing. Given a pixel neighborhood, dilation outputs the pixel with the largest value. Thus, bight objects are enlarged, while dark regions are shrank. The erosion operator performs the dual transformations.

A generalization of the tropical polynomials lies in the notion of hinging hyperplanes by \citeauthor{Wang_Sun_2005} \cite[]{Wang_Sun_2005}, which are a collection of affine terms jointed by points where multiple terms {\it dominate}, i.e. are maximizers or minimizers. These surfaces can be neither convex or concave and correspond to a lattice-theoretic view of these PWL surfaces, also found in the works of \citeauthor{tarela1990representation} \cite{tarela1990representation,tarela1999region}. The maximizers or minimizers are points that are not differentiable and are responsible for the hard edges. In the case of strictly max-plus (or min-plus) PWL surfaces, this collection of points is known as the tropical hypersurface \( \mathcal{ T } (p) \). The hard surfaces can be softened by approximating the \( \min \) and \( \max \) operators, a process known as Maslov Dequantization:

\begin{definition}[Maslov Dequantization \cite{Litvinov_2007}]
    Let \( x,y\in\r\) and \( h>0 \). The transformation \( x\vee_h y=h\log (e^{\nicefrac{x}{h}}+e^{\nicefrac{y}{h}}) \) defines the Maslov Dequantization of the max operator, yielding its soft approximation. Similarly, the Maslov Dequantization of the \( \min \) operator is  \( x\wedge_h y=-h\log (e^{-\nicefrac{x}{h}}+e^{-\nicefrac{y}{h}}) \). As \( h \rightarrow 0  \), the soft versions approach the hard ones: \( \lim_{h \rightarrow 0} x\vee_h y=x\vee y  \) and \( \lim_{h \rightarrow 0} x\wedge_h y=x\wedge y  \) (see proof in Appendix).  
\end{definition}

For small positive values of \( h \), these approximations are part of the Log-Sum-Exp family, used in convex analysis and recently linked to tropical polynomials \cite{Calafiore_Gaubert_Possieri_2018,Calafiore_Gaubert_Member_Possieri_2019}. We use the reciprocal of \( h \), the hardness parameter \( \beta=h^{-1} \). 

	\section{Convex-Concave Procedure Training for General Classification Tasks}  
\label{sec:ccp training}
\subsection{Binary Classification Problem formulation}  

In the neural network training context, the Gradient Descent algorithm and its variants \cite{Ruder_2017} dominate the literature. Network parameters are optimized with stochastic algorithms, allowing parallelization of the process, thus faster training, but sacrificing on robustness. Alternative methods have been proposed that eliminate the stochastic element and offer low variance in the resulting models. A salient example lies in the Support Vector Machines (SVMs) literature, where the training is formulated as a convex program \cite{Boyd_Vandenberghe_2004}, which allows the use of fast and robust algorithms. In the current section, we explore a similar training scheme, based on Difference-of-Convex programming and extend its applicability to multiclass problems.

Let us consider the task of classifying the pattern \( \mathbf{x}_i\in\rn, i=1,2,\dots,N \) to two distinct classes, \( \mathcal{ N }  \) for negative (\( y_i=-1 \)) and \( \mathcal{ P }  \) for positive (\( y_i=+1 \)). Given an input \( \mathbf{x}\in\rn \) and weight vectors \( \mathbf{w,m}\in\r^{n+1} \), a dilation term \( \dilation_\mathbf{w} \) and an erosion term \( \erosion_\mathbf{m} \) are combined to produce an output. The network is shown in Fig. \ref{fig:nn_dep} and carries the name \textit{Dilation-Erosion Perceptron (DEP)}. A similar architecture lies in the maxout network \cite{GWM+13}, which uses the more general notion of tropical polynomials (with arbitrary slopes \( \mathbf{a}_i \), see \( \eqref{eq:tropical polynomial}\)). \citeauthor{Charisopoulos_Maragos_2017} formulate the training of this classifier \cite{Charisopoulos_Maragos_2017} as:

\begin{figure}[]
    \centering
    \includestandalone[width=0.6\linewidth,
    mode=image    
    ]{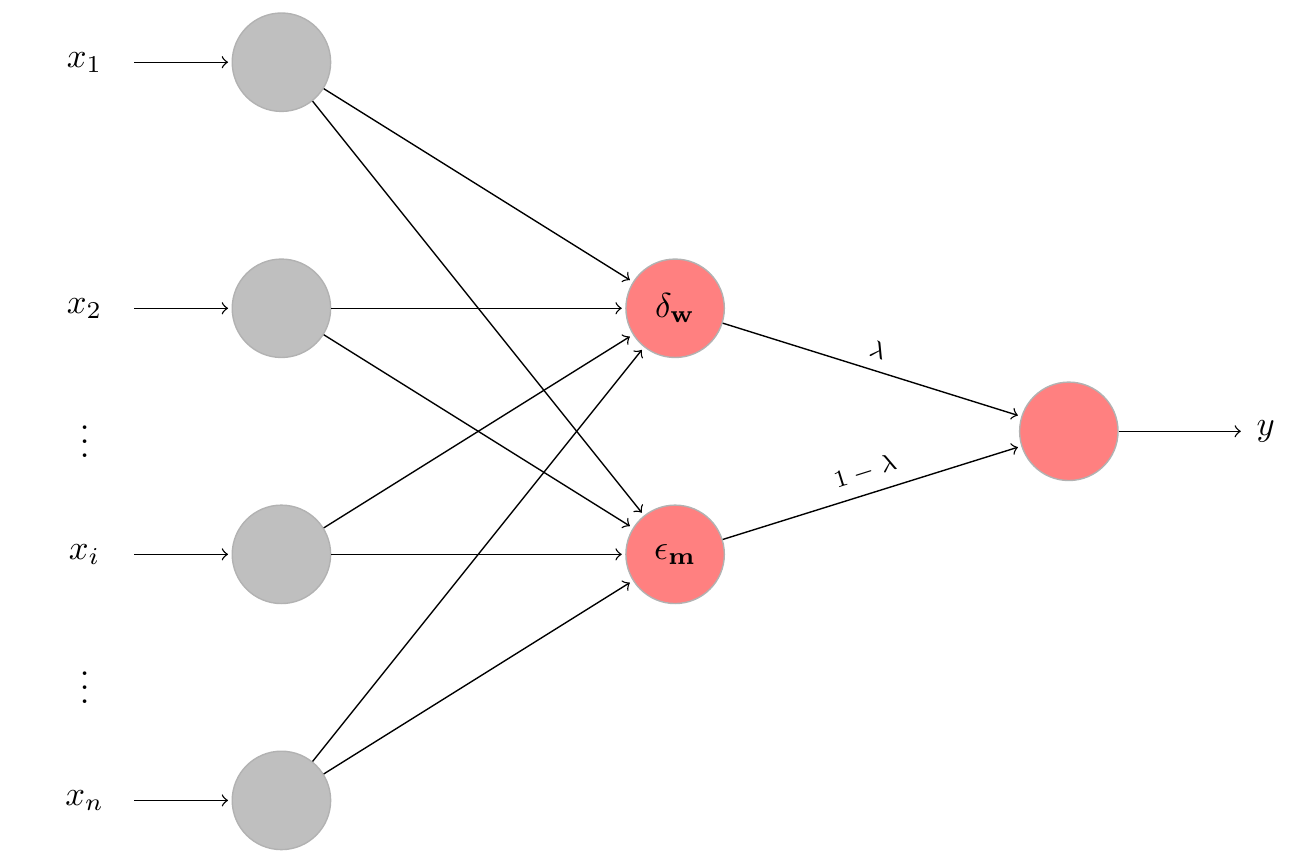}
    \caption{Dilation-Erosion Perceptron (DEP)}
    \label{fig:nn_dep}
\end{figure}
\begin{mini}{}{\sum\limits_{i=1}^{N} v_i\max\{0, \xi_i\} }{}{}
    %  +C \| \mathbf{u-r} \|_1
    \label{eq:ccp dep separate}
    \addConstraint{\lambda \dilation_\mathbf{w}(\x_i)+ (1-\lambda)\erosion_\mathbf{m}(\x_i)}{\geq -\xi_i\quad}{\forall \x_i\in \mathcal{ P } }
    % \addConstraint{f_\mathbf{u}(\x)}{\geq -\xi_i\quad}{\forall \x_i\in \mathcal{ P } }
    \addConstraint{\lambda \dilation_\mathbf{w}(\x_i)+ (1-\lambda)\erosion_\mathbf{m}(\x_i)}{\leq +\xi_i\quad}{\forall \x_i\in \mathcal{ N } }
\end{mini}
where \( \xi_i \) are slack variables and \( v_i \) correspond to a weighting scheme. Specifically, the slack variables \( \xi_i \) ensure that only misclassified patterns are taken into account for the objective function, whereas the weighting scheme allows the attachment of higher importance to some misclassified patterns. \citeauthor{Charisopoulos_Maragos_2017} propose penalizing patterns with greater chances of being outliers. Specifically, for classes \( \mathcal{ C }_0 =\mathcal{ N }  \) and \( \mathcal{ C } _1=\mathcal{ P }  \):
\begin{equation*}
    v_i=\frac{\lambda_i}{\max_j \lambda_j}, \quad \lambda_j = \frac{1}{\| \x_j-\bm{\mu}_k \|_p},\quad \bm{\mu}_k=\frac{1}{|\mathcal{ C } _k|}\sum\limits_{\x_i\in \mathcal{ C } _k} \x_i  
\end{equation*}
The above training formulation \( \eqref{eq:ccp dep separate} \) corresponds to a Difference-of-Convex (DC) optimization problem, since the dilation term is convex whereas the erosion term is concave. This optimization class is vast and includes all \( C^2 \) functions \cite{Hartman_1959}. Various methods have been proposed to tackle such problems with many focusing on the Fenchel Conjugate. Another approach lies in a heuristic called \textit{Convex-Concave Procedure (CCP)}, proposed by \citeauthor[]{Yuille_Rangarajan_2003} \cite{Yuille_Rangarajan_2003} and extended in \cite{Lipp_Boyd_2016,Shen_Diamond_Gu_Boyd_2016}. This heuristic is based on the observation that a DC program is convex iff the concave term, in this case the erosion term, is affine. Hence, by convexifying this term, i.e. calculating its linear approximation via a first-order Taylor series expansion, the problem becomes convex unlocking the ability of efficient and fast solutions. Repeating this process yields the CCP. \citeauthor{Lipp_Boyd_2016} extended this procedure by allowing the initialization of the algorithm without a feasible point and improved convergence with the use of penalty terms \cite{Lipp_Boyd_2016}. CCP was further extended \cite{Shen_Diamond_Gu_Boyd_2016} by the introduction of a structured method of defining DC problems which automatically converts them in a form suitable for generic solvers \cite{Shen_Diamond_Gu_Boyd_2016}. We use such solvers in our experiments, see \texttt{CVXPY} \cite{diamond2016cvxpy}.  

Valle \cite{Valle_2020} proposes a modification of the formulation \( \eqref{sec:ccp training} \), where a greedy algorithm combines the dilation and the erosion perceptrons, which are trained separately beforehand. This method allows the inclusion of a regularization term \( R(\mathbf{u}) \) in the objective function, since one set of parameters is optimized.  Valle proposes \( R(\mathbf{u})= C \| \mathbf{u-r} \|_1 \) , where \( \mathbf{u}=\mathbf{w}\) or \(  \mathbf{u}=\mathbf{m} \) and \( \mathbf{r} \) is a reference term. The linear combination is calculated by minimizing the average hinge loss: 
\begin{equation}
    \label{eq:optimal lambda}
    \lambda^* = \argmin\limits_{0 \leq \lambda \leq 1} \sum\limits_{i=1}^{N} \max\{0,-y_i\left[\lambda \dilation_\mathbf{w}(\x_i)+ (1-\lambda)\erosion_\mathbf{m}(\x_i)\right]\} 
\end{equation}
As a lattice-based model, the Dilation-Erosion Perceptron suffers from a major flaw. Particularly, it presupposes a partial ordering both on the features and the classes. This results in counterintuitive behavior. By simply inverting the classes \( \mathcal{ N } \rightleftarrows \mathcal{ P }   \), the problem intuitively does not change but mathematically it does, which is reflected on a severe drop in performance \cite{Valle_2020}. A method to placate this issue can be found on the use of reduced morphological operators based on a reduced ordering: 
\begin{definition}[reduced ordering]
    Let \( R \) be a nonempty set, \( \mathcal{ L }  \) be a complete lattice and \( \rho:R\rightarrow \mathcal{ L }  \) be a surjective mapping. A reduced ordering, or r-ordering, is defined as:
    \begin{equation}
        \label{eq:r-ordering}
        \x\leq _\rho \mathbf{y} \Leftrightarrow \rho(\x) \leq \rho(\mathbf{y}), \forall \mathbf{x,y}\in R.
    \end{equation}
\end{definition}

Effectively, this means that a mapping \( \rho:\rn \rightarrow \r^m \) which transforms the input is used and the classification is performed on the resulting dataset. Let the original dataset be \( \mathcal{ D } =\{(\mathbf{x}_i,y_i)\in\rn\times\{-1,+1\}: i\in[N  ]\} \), the mapping produces the following dataset \( \mathcal{ D }_\text{new} =\{(\rho(\mathbf{x}_i),y_i)\in\r^m\times\{-1,+1\}: i\in[N  ]\} \) and consists of \( m \)  mappings \( \rho(\mathbf{x})=[\rho_1(\mathbf{x}),\rho_2(\mathbf{x}),\dots,\rho_m(\mathbf{x})]^\top  \) where \( \rho_i(\mathbf{x}):\rn \rightarrow \r,i\in[m] \) are mappings from the original input space to the real numbers and are called \textit{kernels}
. Kernels are extensively used in Support Vector Machine modeling. Some characteristic examples are presented in Table \ref{tab:kernels}. 

\begin{table}[h]
    \centering
    \begin{tabular}{r|l}
        \toprule
        Kernel      & \( k(\mathbf{x,y}) \)\\
        \midrule
        Linear      & \( \left\langle \mathbf{x,y}\right\rangle   \)\\
        Polynomial  & \( \left\langle 1+ \mathbf{x,y}\right\rangle^d \)\\
        Gaussian    & \( e^{-\| \mathbf{x-y} \|^2/(2\sigma^2)} \)\\
        Sigmoid     & \( \tanh(\gamma \left\langle \mathbf{x,y}\right\rangle+r) \)\\
        \bottomrule
    \end{tabular}
    \caption{Kernels}
    \label{tab:kernels}
\end{table}

\subsection{Extension to multiclass problems}  

The Dilation-Erosion Perceptron is inherently a binary classifier. As is the case with Support Vector Machines, this classifier can be extended to tackle multiclass problems. Notably, there are two major approaches: \textit{one-versus-the-rest} and \textit{one-versus-one}. Let us consider a problem with \( K>2 \) classes. On both categories, the idea lies in the construction of several models, all binary, and combine their outputs to produce a single prediction. 

In the former approach, one classifier is trained for each class. In order to do so, the positive class consists of the elements of said class \( \mathcal{ C } _k \) whereas the negative patterns consist of all the other classes \( \mathcal{ C } _{-k} \). A straightforward issue with this approach lies on the imbalance of the datasets. Let \( N \) be the total number of datapoints and consider a relatively uniform distribution of patterns among classes. Then, the positive class is much smaller than the negative: \( |\mathcal{ C } _k | \simeq \frac{N}{K}\ll |\mathcal{ C } _{-k}| \simeq\frac{(K-1)N}{K} \). A way to address this issue is the use of weighting scheme: \( +1 \) for positive and \( -\frac{1}{K} \) for negative patterns. In the latter approach, one classifier is trained for every pair of classes and the prediction is determined by the (hard) majority vote of the corpus of classifiers. The issue that emerges is the sheer number of classifiers that have to be trained. For \( K \) classes, there are \( \frac{K(K-1)}{2} \) pairs. However, each pair is trained on only \( \sim \frac{2}{K}\cdot N \ll N \) patterns.  

In our experiments, we employ the \textit{one-versus-one} approach with a greedy Dilation-Erosion Perceptron \cite{Valle_2020} on the \texttt{MNIST} \cite{lecun1998mnist} and \texttt{FashionMNIST} \cite{xiao2017fashion} datasets. Thus, \( =\frac{10\times 9}{2}=45 \) distinct classifiers must be trained. Each classifier uses a Bagging approach with \( n \) Radial Basis Function (RBF) kernels
%  \footnote{include kernel table??}
 . A Bagging classifier uses the same kernel, in this case RBF, but the parameters are trained in different subsets of the original dataset. The results are presented in Table \ref{tab:rdep:multiclass experiments} and are comparable with gradient-descent methods of training neural networks, presented in the next section. Nevertheless, the networks presented in next sections are deeper and denser and, thus, have many more parameters. An important aspect of this non-stochastic training method is robustness. Repeating the experiments \( 10 \) times yielded very low variance in accuracy, a phenomenon that does not present when training with stochastic gradient descent variants. An important aspect of this method is the kernel selection. Also, there is a different approach other than Bagging, the Ensemble method, where different kernels are used but on the entire training set. Other parameterizations include the number and type of kernels. The combinations are vast and, with further experimentation, the results of Table \ref{tab:rdep:multiclass experiments} can be improved.

\def\nan{\text{NaN}}
\begin{table}[]
    \centering
    \begin{tabular}{l|cc}
        \toprule
        & \texttt{MNIST} & \texttt{FashionMNIST} \\
        \midrule
        \( n=5 \)   & \( \mathbf{97.72\pm 0.01} \) & \( \mathbf{88.21\pm 0.01} \) \\
        \( n=10 \)  & \( \mathbf{97.72\pm 0.01} \) & \( 88.07\pm 0.01 \) \\
        \( n=15 \)  & \( 97.67\pm 0.01 \) & \( 88.11\pm 0.01 \) \\
        \( n=20 \)  & \( 97.64\pm 0.01 \) & \( 88.12\pm 0.01 \) \\
        \bottomrule
        \end{tabular}
    \caption{Results of Bagging \textit{multiclass} reduced DEP  with \( n \) RBF kernels.}
    \label{tab:rdep:multiclass experiments}
\end{table}
	\section{Pruning Morphological Neural Nets}  
\label{sec:sparsity}

\begin{figure}[ht]
    \centering
    \includestandalone[width=0.8\linewidth,mode=image]{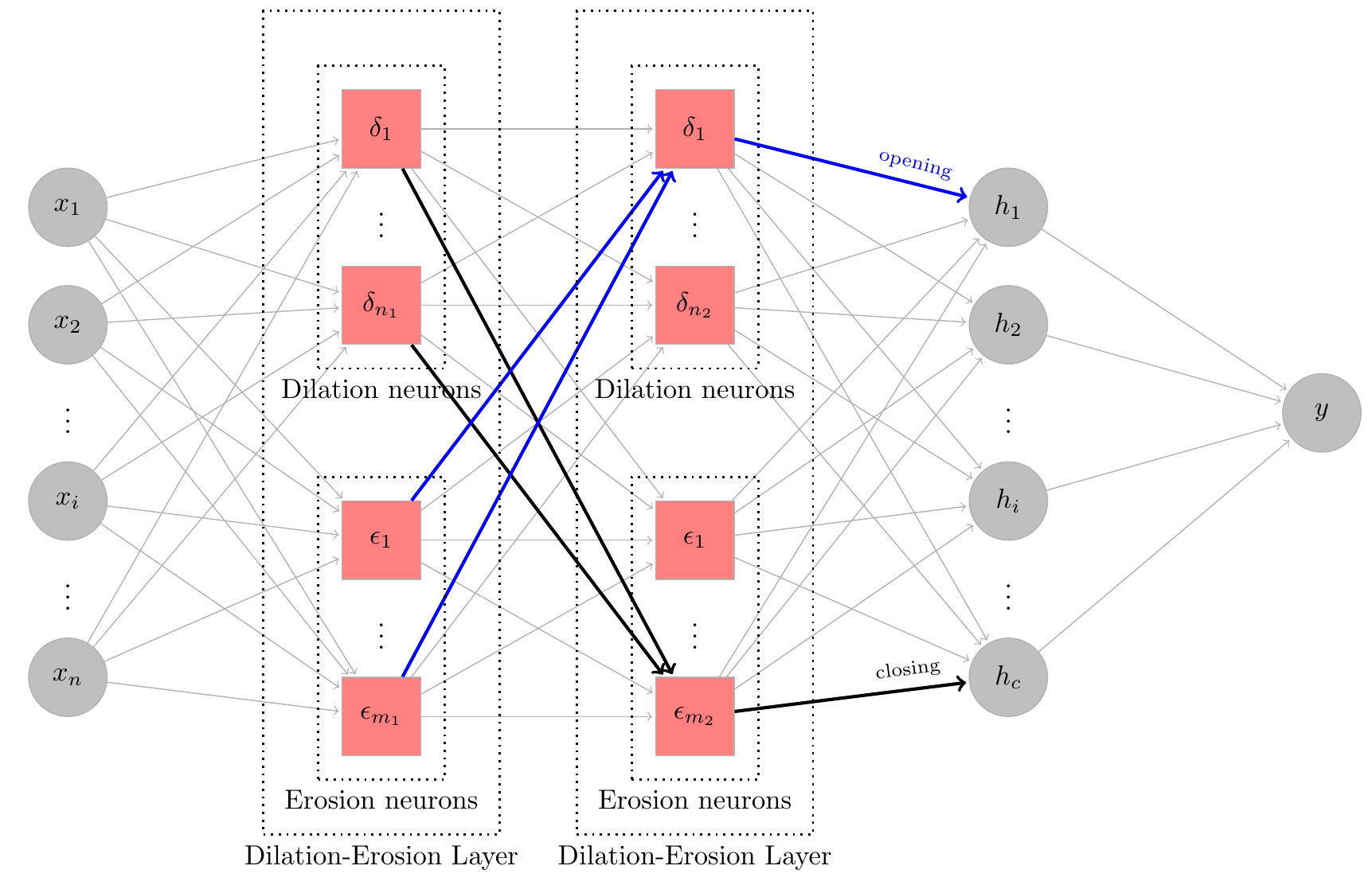}
    \caption{Dense Morphological Network with 2 hidden layers. Square nodes correspond to morphological operators, either \( \min \) (erosion) or \( \max \) (dilation). The output layer is fully connected.}
    \label{fig:denmo 2 hidden}
\end{figure}

The Dilation-Erosion Perceptron has only two neurons in the hidden layer. A straightforward extension is the population of the hidden layer with more neurons and/or the use of multiple hidden layers. This family of networks is called Dense Morphological \cite{Mondal_Santra_Chanda_2019} and an instance is depicted in Fig. \ref{fig:denmo 2 hidden}. The single dilation neuron of DEP is replaced by \( n_1 \) and \( n_2 \) terms in the first and second hidden layer, respectively. Similarly for the erosion neuron. By stacking dilation and erosion neurons on hidden layers, the network calculates activations similar to morphological openings and closings.

By increasing the number of parameters, both in depth and width, the network is able to search for effective hidden representations of the input data, removing the major problem of selecting a surjective mapping \( \rho \), which plagues the multiclass DEP. The use of stochastic optimization methods for training allows for parallelization of the training process and the utilization of optimized deep learning libraries that take advantage of Graphical Process Units (GPUs), resulting in faster training. In this case, the models are trained with standard gradient descent methods. We study two variants: (mini-batch) Stochastic Gradient Descent (SGD) and Adaptive Momentum Estimation (Adam) \cite{Kingma_Ba_2017}. Our focus lies \textit{not} on achieving the highest possible accuracy, but on showing the compression ability of morphological networks compared to traditional ones. To this end, we apply pruning techniques to evaluate the ability of the various networks to retain information with a fraction of the original nodes.

A note on the notation of the models used. Regarding the hidden layer(s), the models studied include: only dilation neurons (denoted as \( \dilation \)), only erosion neurons (\( \erosion \)), a mixed network with both types of neurons (\( (\dilation,\erosion) \)) as well as a Feedforward neural networks with Rectified Linear Unit (ReLU) activations (FF-ReLU) for comparison. The smooth morphological networks are denoted with the subscript \( \beta \). The experiments focus on visual recognition tasks and the datasets selected remain the same as in Section \ref{sec:ccp training}, i.e. \texttt{MNIST} and \texttt{FashionMNIST} . The training lasts \( 50 \) epochs. After experimentation, we conclude that the best learning rates differ for each optimizer and correspond to the values \( \eta=0.001 \) and \( \eta=0.09 \) for Adaptive Momentum Estimation and Stochastic Gradient Descent, respectively. For the Stochastic Gradient Descent, we omit networks with 2 hidden layers and the models with smooth operators, since our experimental setup of 50 epochs does not yield competitive results, due to gradient propagation issues. 

The results are presented in Table \ref{tab:denmo:accuracy comparison}. The Adam algorithm yields higher accuracy for both datasets and both optimizers. Also, the increase in density of the hidden layer is beneficial to accuracy but there is a diminishing returns effect; after a certain amount of nodes in the hidden layer the performance boost is minimal. In both datasets, the linear model FF-ReLU slightly outperforms the morphological equivalents. It is important to note that \citeauthor[]{Mondal_Santra_Chanda_2019} \cite{Mondal_Santra_Chanda_2019} used a different experimental setup (w.r.t. epochs\footnote{\cite{Mondal_Santra_Chanda_2019} uses 400 epochs for training, but we only use 50 epochs. } etc.) and showed that morphological networks can in fact achieve higher accuracy on these datasets compared to FF-ReLU. However, as stated above, our focus lies \textit{not} on accuracy but on compression ability.

{\scriptsize
\begin{table}[h]
    \centering
    \begin{tabular}{cccccccccccccc}
        \toprule
        & {} & \phantom{a}& \multicolumn{6}{c}{Adaptive Momentum Estimation} & & \multicolumn{4}{c}{Stochastic Gradient Descent} \\
        \cmidrule{4-9} \cmidrule{11-14}
        & \( \sharp  \) & &  \( \delta \) & \( \epsilon \) &  \( \delta,\epsilon \) &  \( 2(\delta,\epsilon) \) &  \( \delta_\beta,\epsilon_\beta \) &  FF-ReLU & \phantom{abc}& \( \delta \) & \( \epsilon \) &   \( \delta,\epsilon \) &  FF-ReLU \\
        \midrule
        \multirow{5}{*}{\rotatebox{90}{\scriptsize \texttt{MNIST}}}
        & 32  &       &  94.36 &  87.12 &  92.19 &  92.73 &      92.85   &  96.54 &       &  92.83 &  82.40 &  90.00 &  97.00 \\
        &  64  &       &  96.12 &  91.74 &  94.92 &  95.15 &      95.03 &  97.53 &       &  94.65 &  87.98 &  93.00 &  97.56 \\
        & 128 &       &  96.90 &  94.91 &  96.59 &  96.33 &      96.93 &  98.07 &       &  95.29 &  91.68 &  94.73 &  97.92 \\
        & 256 &       &  97.30  &  94.98 &  97.42 &  97.08 &      96.95 &  98.13 &       & 96.10 &   91.81 &  95.74 &  98.02 \\
        & 400 &       &  97.62 &  96.17 &  97.63 &  97.09 &  97.84 &  98.03 &       &  95.87 &   94.59 &  96.07 &  98.08 \\
        \midrule
        \multirow{5}{*}{\rotatebox{90}{\scriptsize \texttt{FashionMNIST}}}
        & 32  &       &  80.98 &  81.87 &  83.16 &  83.60 &      81.35 &  87.06 &       &  77.75 &  79.43 &  79.82 &  87.49 \\
        & 64  &       &  84.09 &  84.38 &  85.20 &  85.34 &      83.36 &  87.83 &       &  80.41 &  82.44 &  83.10 &  88.29 \\
        & 128 &       &  85.92 &  86.13 &  86.76 &  86.51 &      84.47 &  88.44 &       &  81.21 &  84.18 &  83.98 &  88.66 \\
        & 256 &       &  86.22 &  87.15 &  87.11 &  87.12 &      85.81 &  89.09 &       &      78.65 &  85.77 &  85.25 &  87.69 \\
        & 400 &       &  86.62 &  88.05 &  88.34 &  87.47 &  87.13 &  89.44 &       &      80.09 &  86.20 &  86.21 &  88.81 \\
        \bottomrule
    \end{tabular}
    \caption{Comparison of the accuracy of various network architectures on the \texttt{MNIST} and \texttt{FashionMNIST} datasets. Different sizes of the hidden layers, denoted as \( \sharp \), are evaluated. See above for columns description.}
    \label{tab:denmo:accuracy comparison}
\end{table}
}

\null

The compression ability of the various networks is evaluated by the effect on performance that the removal of a nontrivial amount of parameters has. We apply an \( \ell_1 \)-norm pruning scheme \cite{li2016pruning}, which retains   salient features and removes unimportant ones. By specifying a percentage \( p\in[0,100] \), only the \( p\% \) of the parameters with the highest \( \ell_1 \)-norm are retained. We focus on the hidden layer, where the bulk of the parameters is concentrated. Various percentages of units are pruned and the performance of the remaining network is evaluated on the test set. 

\begin{figure}[ht]
    \def\ll{0.45}
    \centering
    \begin{subfigure}[b]{\ll\linewidth}
        \centering
        \includegraphics[width=0.9\textwidth]{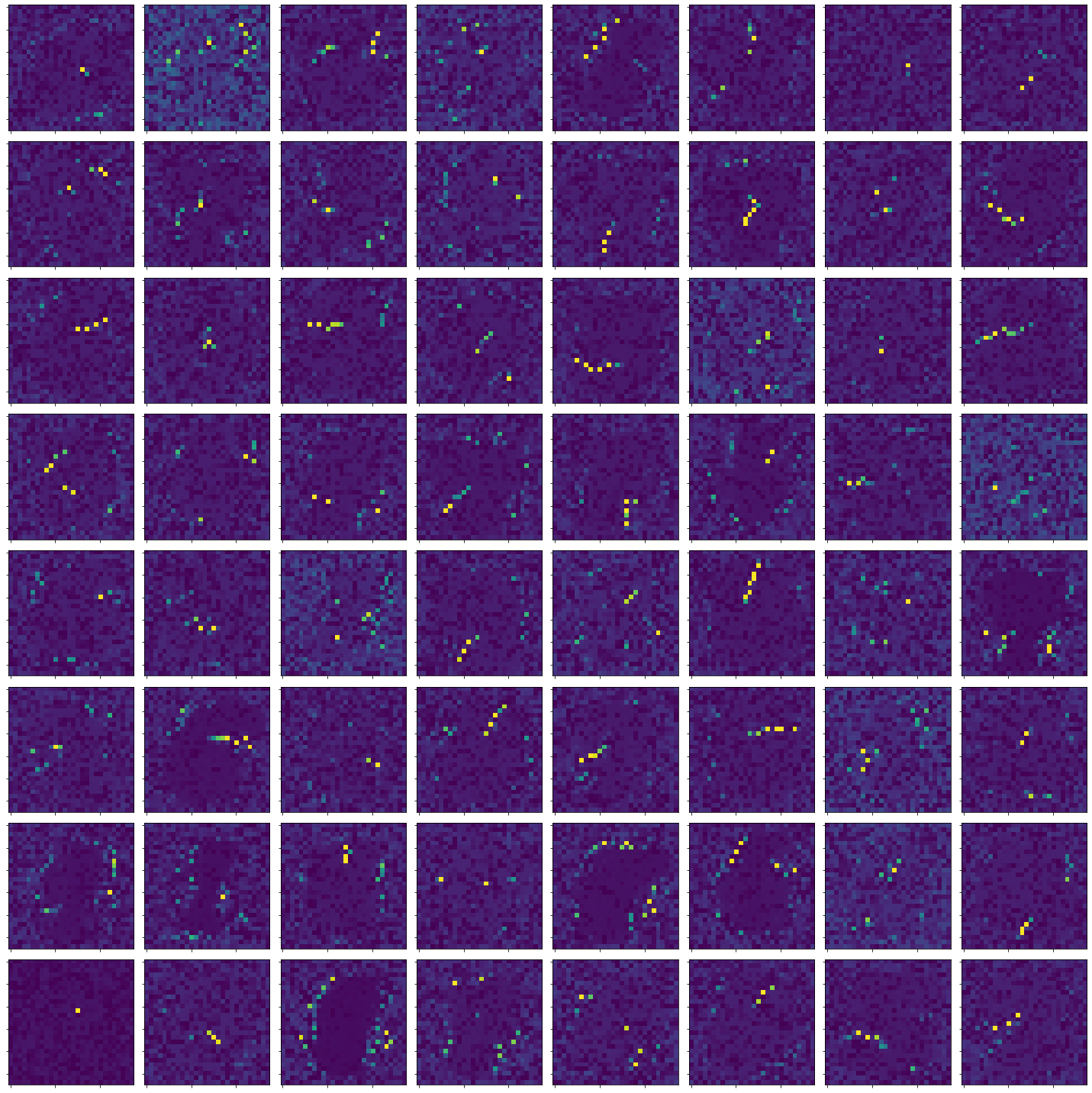}
        \caption{\( (\delta,\epsilon) \)}
        \label{fig:MNIST:active filters:denmo}
    \end{subfigure}
    \hfill
    \begin{subfigure}[b]{\ll\linewidth}
        \centering
        \includegraphics[width=0.9\textwidth]{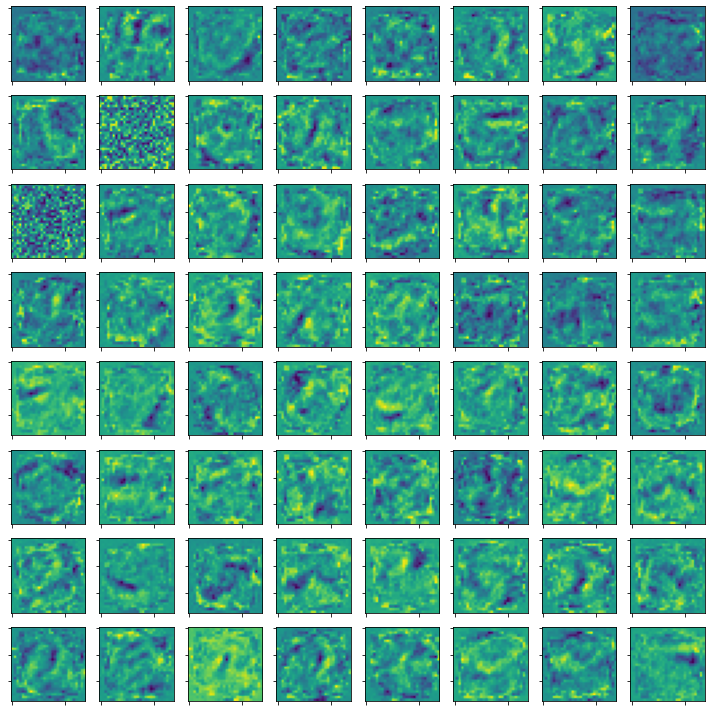}
        \caption{FF-ReLU}
        \label{fig:MNIST:active filters:ff-relu}
    \end{subfigure}
    \caption{Examples of hidden layer activations for various models (\texttt{MNIST} dataset). (a) corresponds to a morphological network (dilation neurons) and (b) to a feedforward network with ReLU activations (FF-ReLU). Both models have \( 64 \) neurons in the hidden layer (forming an \( 8\times 8 \) grid). Each element of the grid has dimensions \( 28\times 28  \) pixels, i.e. equal to the input images of the \texttt{MNIST} dataset.}
    \label{fig:hidden layer activations}
\end{figure}

The results are presented in Table \ref{tab:pruning}. With shades of red, the deterioration in terms of performance is highlighted. All variants of the morphological network significantly outperform their linear counterpart in their ability to retain information and the effectiveness of their constructed hidden representations. This effect becomes more evident in the more complicated dataset \texttt{FashionMNIST}. For example, in the case of the SGD optimizer, the erosion and mixed morphological networks have virtually zero decline in performance (the erosion net has actually a slight increase) even when \( p=1\% \) but the pruning has negatively affected the performance by \( p=25\% \), i.e. FF-ReLU requires \( \sim 25 \) times more parameters in the hidden layer. This difference can be explained by Fig. \ref{fig:hidden layer activations}, where the hidden layer activations are depicted for a morphological and a linear network. Lighter colors correspond to higher values. The representations constructed by the morphological network are sparse; only a fraction of the parameters have high values. Intuitively, this means that the removal of the rest (depicted in blue) will have minimal impact in information loss and, thus, accuracy. However, this effect is not present in the linear net, since the weights have more uniform values. 

\begin{figure}[ht]
    \def\ll{0.22}
    \centering
    \begin{subfigure}[b]{\ll\linewidth}
        \centering
        \includegraphics[width=\linewidth]{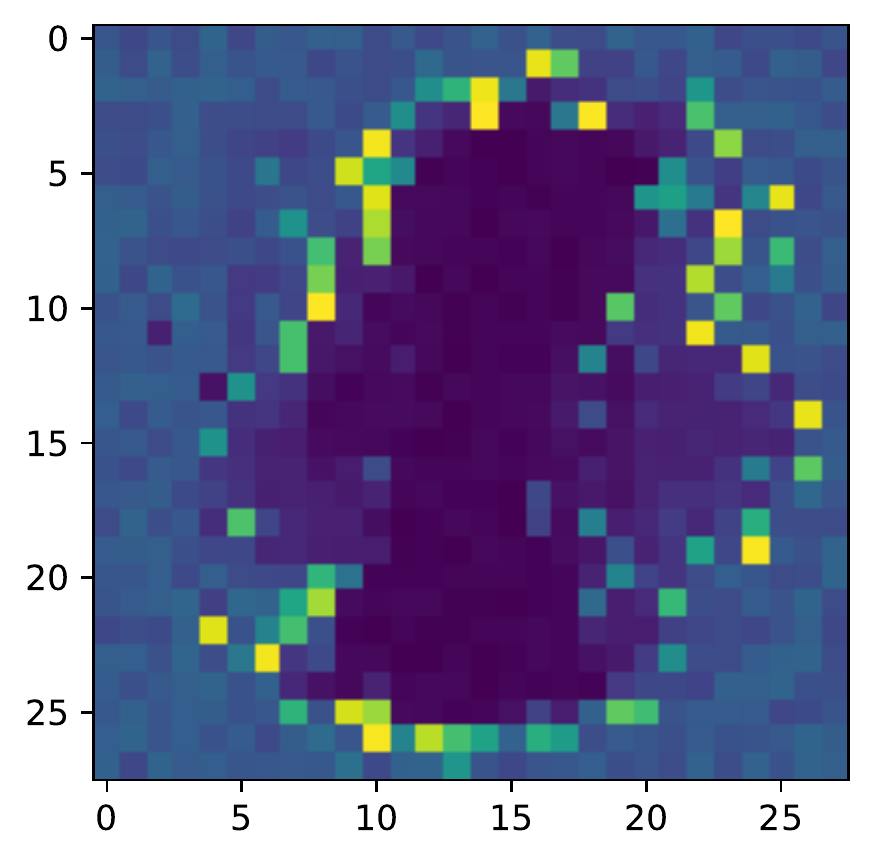}
        \caption{\( (\dilation,\erosion) \) - Adam}
        \label{fig:activations:adam}
    \end{subfigure}
    \hfill
    \begin{subfigure}[b]{\ll\linewidth}
        \centering
        \includegraphics[width=\linewidth]{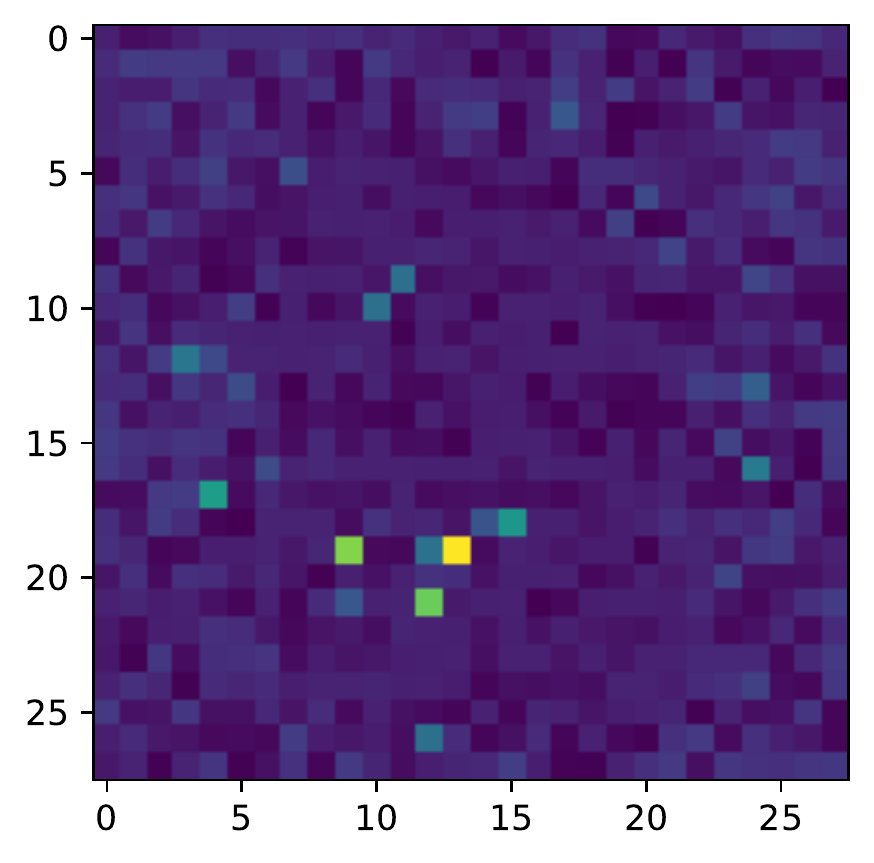}
        \caption{\( (\dilation,\erosion) \) - SGD}
        \label{fig:activations:sgd}
    \end{subfigure}
    % \\
    \hfill
    \begin{subfigure}[b]{\ll\linewidth}
        \centering
        \includegraphics[width=\linewidth]{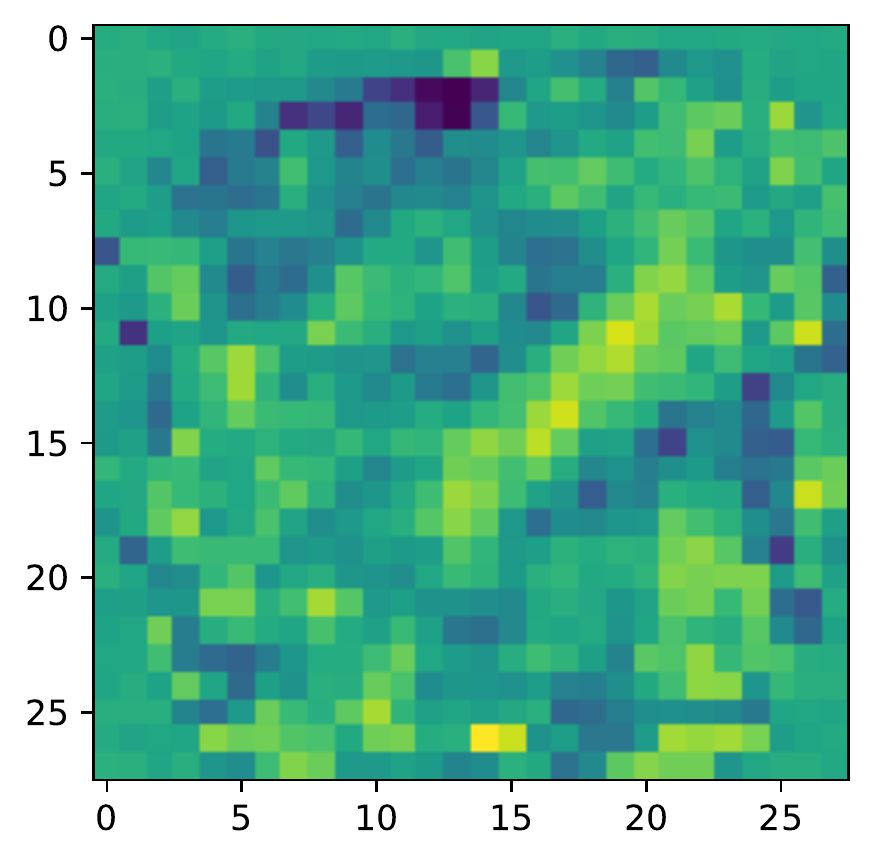}
        \caption{FF-ReLU - Adam}
        \label{fig:activations:ff-relu:adam}
    \end{subfigure}
    \hfill
    \begin{subfigure}[b]{\ll\linewidth}
        \centering
        \includegraphics[width=\linewidth]{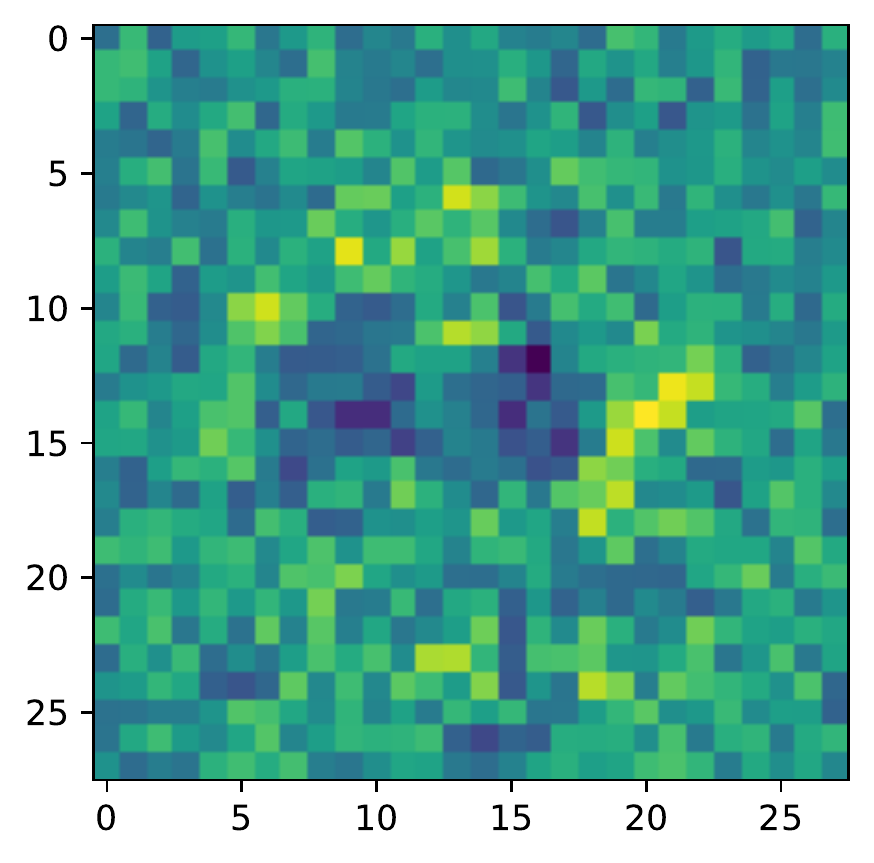}
        \caption{FF-ReLU - SGD}
        \label{fig:activations:ff-relu:sgd}
    \end{subfigure}
    \caption{Examples of hidden layer activations for various models (\texttt{MNIST} dataset). (a) and (b) correspond to a morphological network (dilation neurons), whereas (c) and (d) correspond to a feedforward network with ReLU activations.}
    \label{fig:activations}
\end{figure}

Another insight lies in the optimizer selection. Even though models trained with Stochastic Gradient Descent have lower accuracy scores, their ability to retain information is superior than those trained via Adaptive Momentum Estimation. Quantitatively, the optimizer effect can be seen in Table \ref{tab:pruning} by comparing the first and last rows (\( p=100\% \) vs \( p=1\% \)). The SGD models have no loss in accuracy, whereas the Adam models have a slight decrease. Quantitatively, this effect can be explained by Fig. \ref{fig:activations}, where a single neuron of networks with 400 neurons in the hidden layer is depicted for each model category. In the case of the Adam-trained morphological network, the activation is reminiscent of the the outline of a \texttt{MNIST} pattern and uses a nontrivial amount of parameters. On the other hand, Fig. \ref{fig:activations:sgd} shows notable sparsity in the SGD-trained model activations. The effect holds for the linear networks but is milder given the already dense activations.

In conclusion, the morphological networks are characterized by extreme economy in their constructed representations. This is due to the fact that a morphological neuron, dilation or erosion, allows only one element \( x_i \) of the input \( \mathbf{x}\in\rn \) to determine its output. With a small (to none) decrease in accuracy compared to their linear counterparts, the morphological networks are able to retain their performance with a small fraction of the original parameters and display higher compression ability.

\begin{table*}[ht]
    \centering
    
    \begin{tabular}{crcccccccccccccccc}
        % \begin{tabular}{cr|c|ccccccc|c|ccccccc}
        \toprule
        & {} & \phantom{a}& \multicolumn{4}{c}{Adaptive Momentum Estimation} &  & \multicolumn{4}{c}{Stochastic Gradient Descent} \\
        \cmidrule{4-7} \cmidrule{9-12}
        & \( p \) & &  \( \dilation \) & \( \erosion \) &  \( (\dilation,\erosion) \) &  FF-ReLU  & \phantom{abc}& \( \dilation \) & \( \erosion \) &  \( (\dilation,\erosion) \) &  FF-ReLU  \\
        % \midrule
        % & \\
        \midrule
        \multirow{9}{*}{\rotatebox{90}{\texttt{MNIST}}}
        % \bottomrule
        & 100\% &      &  97.62 &  96.17 &  97.95 &  98.13 &     &  94.86 &  93.36 &  96.07 &  98.16 \\
        & 75\%  &      &  97.62 &  96.18 &  97.93 &  98.15 &     &  94.86 &  93.36 &  96.07 &  98.12 \\
        & 50\%  &      &  97.62 &  96.22 &  97.90 &  98.17 &     &  94.86 &  93.37 &  96.07 &  98.08 \\
        & 25\%  &      &  97.62 &  96.09 &  97.87 &  97.51 &     &  94.86 &  93.40 &  96.06 &  98.01 \\
        & 10\%  &      &  97.62 &  95.78 &  97.74 &  \bbmm93.38 &     &  94.86 &  93.38 &  96.09 &  \bbmm96.67 \\
        & 7.5\% &      &  97.62 &  95.42 &  97.76 &  \bb 90.17 &     &  94.86 &  93.38 &  96.10 &   \bbmm95.56 \\
        & 5\%   &      &  97.62 &  \bbmm94.51 &  97.66 &  \bb83.39 &     &  94.86 &  93.40 &  96.10 &  \bb 92.96 \\
        & 2.5\% &      &  97.62 &   \bbmm93.43 &  97.37 &  \bb68.93 &     &  94.86 &  93.39 &  96.09 &  \bb 80.48 \\
        & 1\%   &      & \g 97.62 &   \bbmm91.17 &  97.08 &  \bb44.22 &     &  \g 94.86 & \g 93.38 & \g 96.08 &  \bb 58.07 \\
        % [1em]
        \midrule
        % & \\
        \multirow{9}{*}{\rotatebox{90}{\texttt{FashionMNIST}}}
        & 100\% &      &  86.31 &  86.82 &  88.32 &  88.82 &     &  82.06 &  85.23 &  86.21 &  87.79 \\
        & 75\%  &      &  86.30 &  86.81 &  88.30 &  88.88 &     &  82.00 &  85.23 &  86.21 &  87.75 \\
        & 50\%  &      &  86.22 &  86.80 &  88.33 &  88.18 &     &  82.05 &  85.25 &  86.20 &  87.19 \\
        & 25\%  &      &  85.95 &  86.85 &  88.31 &  \bbmm82.15 &     &  81.90 &  85.26 &  86.28 &  \bbmm 84.35 \\
        & 10\%  &      &  85.58 &  86.27 &  88.05 &  \bb 65.89 &     &  81.67 &  85.27 &  86.23 &  \bb 73.22 \\
        & 7.5\% &      &  85.47 &  86.15 &  87.99 &  \bb 57.93 &     &  81.63 &  85.27 &  86.21 &  \bb 63.95 \\
        & 5\%   &      &  85.37 &  85.81 &  87.76 &  \bb 49.12 &     &  81.52 &  85.24 &  86.22 &  \bb 47.73 \\
        & 2.5\% &      &  84.91 &  85.47 &  87.56 &  \bb 42.48 &     &  81.14 &  85.26 &  86.22 &  \bb 38.84 \\
        & 1\%   &      & \bbmm 81.14 & \bbmm 84.86 &  \bbmm86.85 &  \bb 28.13 &     &  80.68 & \g 85.27 &  \g86.18 &  \bb 35.46 \\
        \bottomrule
    \end{tabular}
    \caption{Performance of pruned networks on the \texttt{MNIST} and \texttt{FashionMNIST} datasets for various model architectures. With shades of red, we show the rapid deterioration in performance. With green, we draw attention to the \textit{absence} of loss in accuracy performance between the full (unpruned) network ant the pruned network with only \( 1\% \) of the parameters in the hidden layer.}
    \label{tab:pruning}
\end{table*}

	\section{Monotonicity Constraints}  
\label{sec:monotonicity}

\begin{figure}[ht]
    \centering
    \includestandalone[width=0.8\linewidth,mode=image]{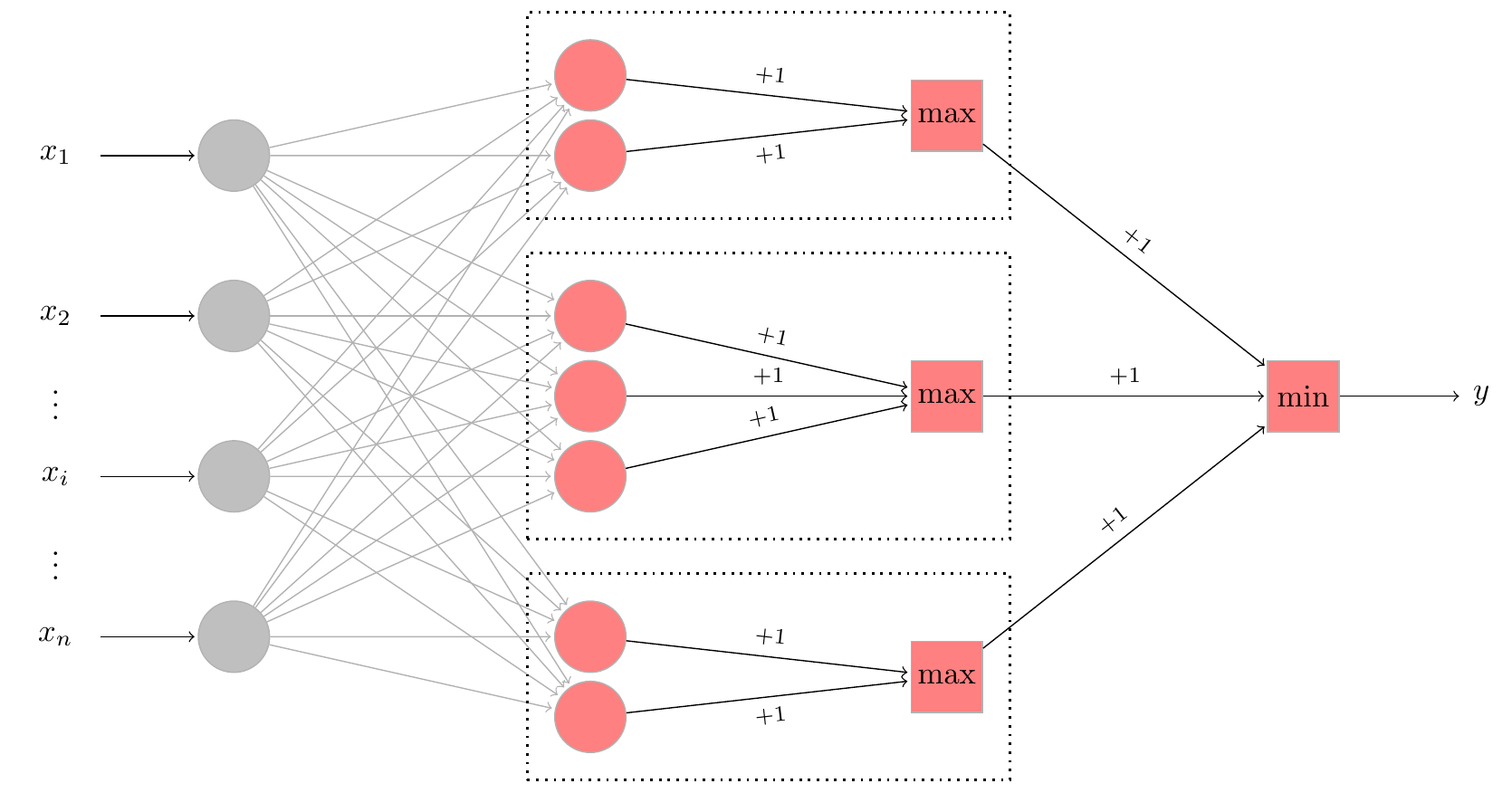}
    \caption{Monotonic network. The gray edges correspond to nonnegative weights.}
    \label{fig:sill}
\end{figure}

In this section, we explore how architectural structure can be used to enforce shape constraints. We examine the case of monotonicity. The output of a network is called monotone if it does not decrease with the increase of the input\footnote{Without loss of generality, we consider only monotonically increasing functions.} The network into consideration was introduced by Sill \cite{Sill} and is depicted in Fig. \ref{fig:sill}. Max-affine terms, i.e. dilations or max-plus polynomials, are succeeded by a min-pooling layer, i.e. an erosion term or a min-plus polynomial. Sill names as a \textit{group} each of the \( K \) max terms, each of which comprises of \( J_k \) affine terms or hyperplanes, \( k=1,2,\dots,K \). For simplicity, we consider the case where \( J_k=J \), \( \forall k=1,2,\dots,K \). Then, for input \( \x \in\rn \), the output is:
\begin{equation}
    \label{eq:sill:output}
    y = f(\x) = \bigwedge\limits_{k\in[K]}\bigvee\limits_{j\in[J]} \{\mathbf{w}_{k,j}^\top \x + b_{k,j}\}
\end{equation} 
Monotonicity constraints are enforced by limiting the weight vector \(  \mathbf{w}  \) to nonnegative values  via a function \( f \) with a positive image \( f:\rn\rightarrow \rn_{\geq 0} \). Various such functions have been proposed. Sill used an an exponential transformation \( w_i = e^{z_i}, z_i\in\r \), whilst \citeauthor[]{Velikova_Daniels_Feelders_2006} \cite{Velikova_Daniels_Feelders_2006} propose the transformation \( w_i=z_i^2 \), which suffers no arithmetic issues and offers the possibility of flat surfaces when \( \mathbf{z}=\mathbf{0} \). The network produces a PWL approximation of the input data. Max-plus terms construct convex surfaces, whereas min-plus terms construct concave surfaces. Their combination in this network structure accommodates for the more general nature of a monotonic surface, which combines both convex and concave parts. The resulting surface can approximate any monotone function to an arbitrary degree of accuracy \cite[Theorem 3.1]{Sill}. 

From a mathematical morphology viewpoint, the application of dilation succeeded by an erosion term results in a closing. For strictly positive weights, the transformation is reversible and corresponds to a morphological opening \( x = f^{-1}(y)=\bigvee_{k\in[K]}\bigwedge_{j\in[J]} \{w_{k,j}^{-1} (y - b_{k,j})\} \) \cite{Dutting_Feng_Narasimham_Parkes_Ravindranath}. An opening \( \alpha=\delta\epsilon \) is an increasing, idempotent and anti-extensive operator, while a closing \( \beta=\epsilon\delta \) shares the first two properties but is extensive. Opening and closing form an adjunction pair \cite{Maragos_2017}.

The morphological terms are not differentiable, which has immediate effect in training via stochastic gradient descent variants, since the backpropagation step does not have an analytical form. Subgradient methods can be used in this case. The backpropagation equations for the dilation and erosion terms of Sill's monotonic network are:

\begingroup
\allowdisplaybreaks
\begin{align*}
    \label{eq:sill:backprop:dilation}
    \frac{\partial \epsilon}{\partial \delta_k} &=\begin{cases}
        1 & \argmax_h\{\delta_h\} = k\\
        0 & \text{otherwise}
    \end{cases}
    \numberthis
    \\
    \label{eq:sill:backprop:weights}
    \frac{\partial \delta_k}{\partial \mathbf{w}_{k,j}} &=\begin{cases}
        \mathbf{x} & \argmax_h\{\mathbf{w}_{k,h}^\top \mathbf{x}\} = j\\
        0 & \text{otherwise}
    \end{cases}
    \numberthis
\end{align*}
\endgroup

In other words, the morphological operators are selective in the sense that a single element \( x_i \) of the input vector \( \x \) is solely responsible for the output. Thus, in the backpropagation step, only \( x_i \)'s parameters are updated. The effect is intensified in Sill's network where two morphological layers are stacked. This implies that only the active hyperplane's weights are updated for a given pattern. Out of the \( K\times J \) affine terms, only a single one gets updated. This results in slow convergence and, potentially, poor approximations. Given low initialized weight parameters, the updated hyperplanes have acquired higher values in order to approximate the data and, subsequently, dominate the groups, not allowing the remaining (hyperplanes) to update their weights. 

Sill proposes a variant of the gradient descent algorithm where the gradient for each hyperplane is computed by the error of the patterns corresponding to the active hyperplane at each iteration. Another method to address this issue is the use of a gain parameter \( G \) which magnifies the initialized weights. This way, the weights of the dominating hyperplane get diminished during the backpropagation step allowing the other hyperplanes to dominate in the next epochs. We propose a method to circumvent this issue completely, which lies in the softening of morphological operators  \( \max \) and \( \min \) via Maslov Dequantization, alleviating the undifferentiability of their hard counterparts.

\begin{figure}[h]
    \centering
    \includegraphics[width=0.8\linewidth]{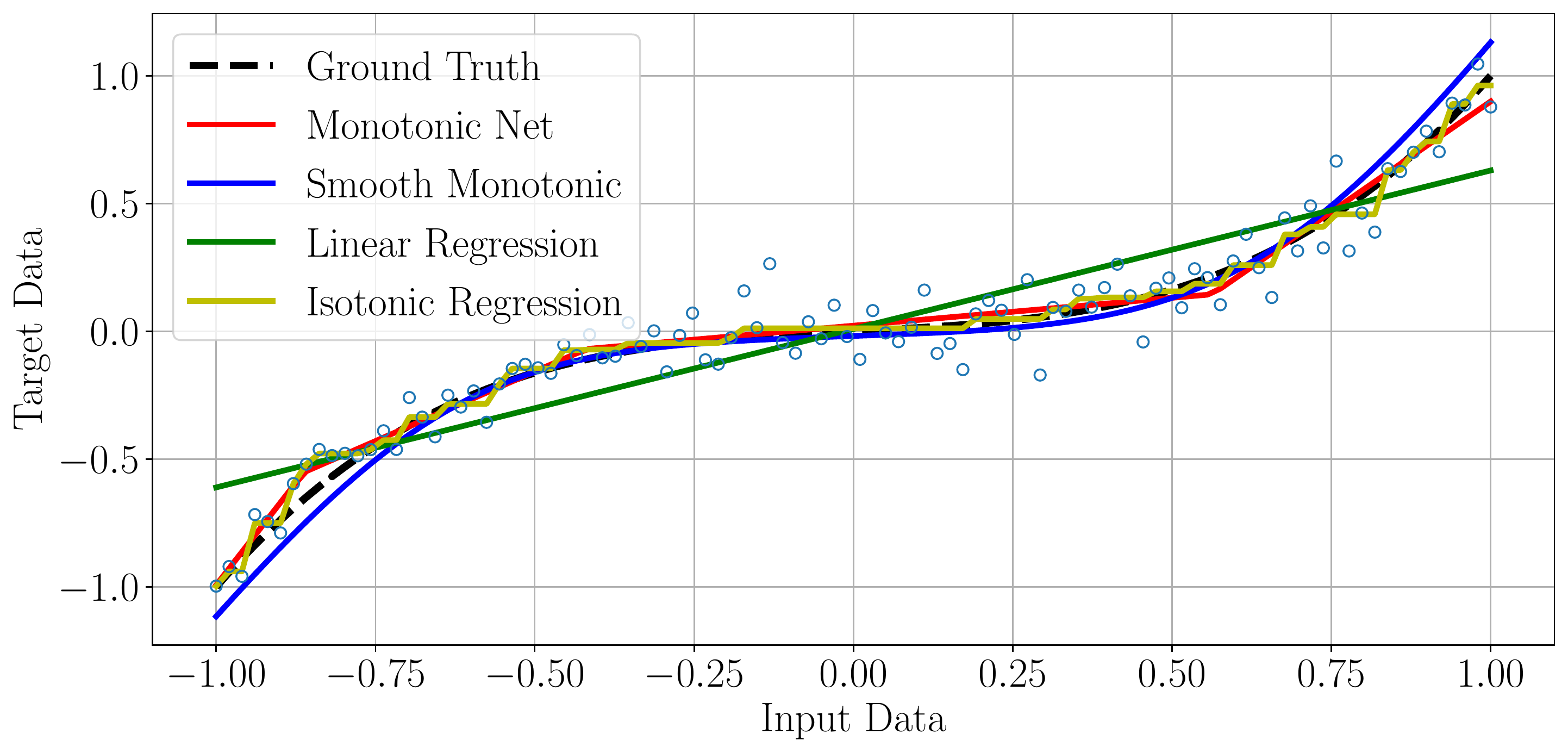}
    \caption{Comparison of monotonic regression methods}
    \label{fig:monotonic outputs}
\end{figure}

We illustrate this method via a simple example. We consider the (strictly) increasing function \( f(x)= x^3 + x +\sin x, x\in[-4,4]\) and scale both domain and image to \( [-1,1] \). Glorot uniform initialization \cite{glorot2010understanding} is used for all network weights. 100 observations are sampled uniformly and corrupted with additive i.i.d zero-mean Gaussian noise \( \epsilon\sim \mathcal{ N } (0,\sigma^2) \). The training lasts 1000 epochs using Adaptive Momentum Estimation with learning rate \( \eta=0.01 \). For comparison purposes, we use isotonic regression \cite{barlow1972isotonic}, a method based on quadratic programming that yields monotone piecewise constant surfaces and has the following formulation in \( \r \):
\begin{mini*}{}{\sum_i w_i(y_i-\hat{y}_i)^2}{}{}
    \addConstraint{\hat{y_i}<\hat{y_j}}{\qquad x_i<x_j }
\end{mini*}
For the monotonic net, we select gain parameter \( G=20 \) for initializing the weights. For the smooth monotonic net, we select hardness parameter \( \beta=5 \). The results are presented in Table \ref{tab:monotonic net} for \( K=J=5 \) for various noise levels. Particularly, for \( \sigma=0.15 \) the surfaces that each method produces are depicted in Fig. \ref{fig:monotonic outputs}. 
From Table \ref{tab:monotonic net}, we conclude that the smooth monotonic net outperforms the other methods for all noise levels \( \sigma \). Furthermore, its training procedure needs not the selection of an arbitrary gain parameter \( G \) and is overall simpler and more intuitive.

\begin{table}[h]
    \centering
    \begin{tabular}{r|llll}
        \toprule
        \( \sigma \)  & \( 0.05 \) & \( 0.1 \)& \( 0.15 \) & \( 0.2 \) \\
        \midrule
        Linear Reg.      & $0.0236$ & $0.03077$ & $0.04827$& $0.0505$\\
        Isotonic Reg.    & $0.0042$ & $0.01112$ & $0.02557$& $0.0417$ \\
        Sill Net          & $0.00305$& $0.01107$ & $0.02401$& $0.0390$\\
        Smooth Sill Net   & $\mathbf{0.00294}$ &$\mathbf{0.00938}$ & $\mathbf{0.02302}$& $\mathbf{0.0386}$\\
        \bottomrule
    \end{tabular}
    \caption{RMS error of monotonic regression methods with noise  \(\erosion\sim \mathcal{ N } (0,\sigma^2) \)}
    \label{tab:monotonic net}
\end{table}

	\section{CONCLUSION}   
\label{sec:conclusion} 

In this paper, we have proposed extensions of morphological neurons regarding their training, enforcing shape constraints such as monotonicity via architectural choices and studied their compression ability. We extended the binary classifier \textit{Dilation-Erosion Perceptron} to general classification tasks using the \textit{one-versus-one} approach and employed a Bagging method with Radial Basis Functions kernels to alleviate the partial ordering of lattice-based models and developed robust classifiers with accuracies comparable to more general morphological networks. We studied the accuracy of dense morphological networks under heavy pruning and compared their ability to construct efficient hidden representations to their linear counterparts, drawing favorable conclusions for the morphological models. Finally, we proposed the use of smooth operators in a monotonic network which improves not only performance but convergence in training as well.

	\printbibliography
	\clearpage\null
	\appendix
\section{Appendix}  

\begin{proof}
    We will prove that \( \delta_{\beta}(\mathbf{x}) = \frac{1}{\beta}\log\left(\sum\limits_{k} e^{\beta x_k} \right) \rightarrow \max_k\{x_k\} = \delta(\mathbf{x} )\) as \( \beta\rightarrow \infty  \). 
    \begin{align*}
        % \label{eq:<label>}
        \lim\limits_{\beta \rightarrow \infty}  \delta_{\beta}(\mathbf{x})
            &=\lim\limits_{\beta \rightarrow \infty} \frac{1}{\beta}\log\left(\sum\limits_{k} e^{\beta y_k} \right)\\
            &=\lim\limits_{\beta \rightarrow \infty} \frac{\log\left(\sum\limits_{k} e^{\beta y_k} \right)}{\beta} \tagg{DLH}\\
            &=\lim\limits_{\beta \rightarrow \infty} \frac{\sum\limits_{k} y_k e^{\beta y_k}}{\sum\limits_{k} e^{\beta y_k}}\\
            &=\lim\limits_{\beta \rightarrow \infty} \sum\limits_{j} \frac{ y_j e^{\beta y_j}}{\sum\limits_{k} e^{\beta y_k}}\\
            &= \lim\limits_{\beta \rightarrow \infty} \sum\limits_{j} \frac{ y_j }{1+ \sum\limits_{k\neq j} e^{\beta (y_k-y_j)}} \tagg{divide by \( e^{\beta y_j} \)} \\
            &=\sum\limits_{j} \underbrace{\lim\limits_{\beta \rightarrow \infty} \frac{ y_j }{1+ \sum\limits_{k\neq j} e^{\beta (y_k-y_j)}} }_{A_j}
            \label{eq:proof:soft maximum:step 1}\numberthis
    \end{align*}  
    where with "DLH" we note the use of the \( De \text{  }  L'H\hat{o}pital  \) rule. Let \( K \) be the number of elements that are equal to the maximum term. Then, we have the following cases for the term \( A_j \): 
    \begin{itemize}
        \item \( j \neq j^* =  \argmax\limits_k y_k \). Then, there exist three sets \( I^>, I^<, I^= \) such that \( I^> = \{i: y_i > y_j \}  \), \( I^< = \{i: y_i < y_j \}  \) and \( I^= = \{i: y_i = y_j \}  \). But \( I^>\neq \emptyset \). Hence, 
        \begin{align*}
            % \label{eq:proof:soft maximum:step 2}
            A_j &= \lim\limits_{\beta \rightarrow \infty} \frac{ y_j }{1+ \sum\limits_{k\neq j} e^{\beta (y_k-y_j)}} 
            \\&= \lim\limits_{\beta \rightarrow \infty} \frac{ y_j }{1+ \sum\limits_{k \in I^<} e^{\beta (y_k-y_j)} + \sum\limits_{k \in I^=} e^{\beta (y_k-y_j)}+  \sum\limits_{k \in I^>} e^{\beta (y_k-y_j)} } 
            \\&= \frac{ y_j }{1+ \sum\limits_{k \in I^<} e^{-\infty } + \sum\limits_{k \in I^>} e^0+ \sum\limits_{k \in I^>} e^{\infty } } = 0
        \end{align*}
        \item \( j = j^* = \argmax\limits_k y_k  \). Then, \( I^> = \emptyset, |I^=|=K-1    \) from our hypothesis and each term \( y_k-y_j \) of the exponential at the denominator is negative for all \( k\in I^< \). Thus:
        \begin{align*}
            % \label{eq:proof:soft maximum:step 2}
            A_j &= \lim\limits_{\beta \rightarrow \infty} \frac{ y_j }{1+ \sum\limits_{k\neq j} e^{\beta (y_k-y_j)}} 
            \\&= \frac{ y_j }{1+\sum\limits_{k \in I^=} e^0+  \sum\limits_{k \in I^<} e^{-\infty } }
            \\&= \frac{ y_j }{1+ (K-1) +  0 } 
            \\&= \frac{y_j}{K}=\frac{y_{\max}}{K}
        \end{align*}
    \end{itemize}
    Subsequently, the quantity \( A_j \) is zero for every \( y_j \) that is not equal to the maximum \( y_{\max} \) and is equal to \( \frac{y_{\max}}{K} \) for each of the \( K \) terms that are equal to the maximum. Thus, using \ref{eq:proof:soft maximum:step 1} concludes the proof.
\end{proof}
\end{document}